\documentclass[conference]{IEEEtran}
\IEEEoverridecommandlockouts
\usepackage{cite}
\usepackage{amsmath,amssymb,amsfonts}
\usepackage{algorithmic}
\usepackage{graphicx}
\usepackage{textcomp}
\usepackage{xcolor}

\usepackage{color, colortbl}
\usepackage{lipsum}
\usepackage{xspace}
\usepackage{booktabs}
\usepackage[utf8]{inputenc}
\usepackage{pifont}

\usepackage[ruled,vlined,linesnumbered]{algorithm2e}

\def\BibTeX{{\rm B\kern-.05em{\sc i\kern-.025em b}\kern-.08em
    T\kern-.1667em\lower.7ex\hbox{E}\kern-.125emX}}

\definecolor{Highlight}{HTML}{39b54a}  
\definecolor{HighlightR}{HTML}{ed2939}  

\newcommand{\cgaphl}[2]{
\fontsize{7pt}{1em}\selectfont{\textcolor{Highlight}{(${#1}$\textbf{#2})}}
}
\newcommand{\cgapg}[2]{
\fontsize{7pt}{1em}\selectfont{\textcolor{Highlight}{(${#1}$#2)}}
}

\definecolor{Gray}{gray}{0.9}

\SetKwInput{KwFunc}{Funcs}

\newcommand{\sage}{\textcolor{black}{\texttt{SAGE-NDVI}}\xspace}

\DeclareRobustCommand*{\IEEEauthorrefmark}[1]{%
  \raisebox{0pt}[0pt][0pt]{\textsuperscript{\footnotesize #1}}%
}

\begin{document}

\title{\sage: A Stereotype-Breaking Evaluation Metric for Remote Sensing Image Dehazing Using Satellite-to-Ground NDVI Knowledge
}

\author{
	\IEEEauthorblockN{
	Zepeng Liu\IEEEauthorrefmark{1,2}$^{,\dagger}$\thanks{$^{\dagger}$This work was done when Zepeng Liu and Mingye Zhu were interns at Ping An Technology, Shenzhen, Guangdong, China, and when Yibing Wei was an intern at PAII Inc., Palo Alto, CA, USA.}, 
	Zhicheng Yang\IEEEauthorrefmark{3}$^{,\ddagger}$\thanks{$^\ddagger$Corresponding authors: Zhicheng Yang ($\mathtt{zcyangpingan@gmail.com}$); Jun Yu ($\mathtt{harryjun@ustc.edu.cn}$); Jui-Hsin Lai ($\mathtt{juihsin.lai@gmail.com}$)}, 
	Mingye Zhu\IEEEauthorrefmark{1,2}$^{,\dagger}$, 
	Andy Wong\IEEEauthorrefmark{3},
	Yibing Wei\IEEEauthorrefmark{4,3}$^{,\dagger}$,
	Mei Han\IEEEauthorrefmark{3}, \\
	Jun Yu\IEEEauthorrefmark{1}$^{,\ddagger}$, 
	Jui-Hsin Lai\IEEEauthorrefmark{3}$^{,\ddagger}$
	}
        \IEEEauthorblockA{
		\IEEEauthorrefmark{1}University of Science and Technology of China, China\\
        \IEEEauthorrefmark{2}Ping An Technology, China\\
		\IEEEauthorrefmark{3}PAII Inc., USA\\
		\IEEEauthorrefmark{4}University of Wisconsin - Madison, USA
	}
}

\maketitle

\begin{abstract}
Image dehazing is a meaningful low-level computer vision task and can be applied to a variety of contexts. In our industrial deployment scenario based on remote sensing (RS) images, the quality of image dehazing directly affects the grade of our crop identification and growth monitoring products. However, the widely used peak signal-to-noise ratio (PSNR) and structural similarity index (SSIM) provide ambiguous visual interpretation. In this paper, we design a new objective metric for RS image dehazing evaluation. Our proposed metric leverages a ground-based phenology observation resource to calculate the vegetation index error between RS and ground images at a hazy date. Extensive experiments validate that our metric appropriately evaluates different dehazing models and is in line with human visual perception.
\end{abstract}


\section{Introduction}
\label{sec:intro}

As an integral part of low-level computer vision (CV) tasks, image dehazing is utilized to remove the influence of weather factors and improve the visual effects of the images. Image dehazing can be applied to a wide range of scenarios, including game production and vision systems for autonomous driving, video surveillance, military reconnaissance, etc.
Another major dehazing scenario is in the field of \emph{remote sensing} (RS) imagery, with both government-level and commercial-level applications, such as change detection and crop identification.

In our industrial scenario, our customers' demands are crop identification and growth monitoring using RS images \cite{du2023parcs,huang2022unsupervised,wong2023knowledge}. These demands heavily rely on the calculation of the Normalized Difference Vegetation Index (NDVI) \cite{rouse1974monitoring}, a widely used RS index to assess vegetation growth. However, RS images with fog and haze can cause significant errors in NDVI calculation, leading to the unsatisfactory results we deliver to our customers. Therefore, image dehazing assessment is essential to convey the proper RS image product to our clients.

In order to evaluate the dehazing effects, a dehazed image and a clear ground truth image are required to estimate their similarity. This can be easily achieved in the natural image domain, because many dehazing datasets employ clear images to artificially generate the corresponding hazy images. However in the RS domain, a hazy image does not have a clear counterpart at the same time stamp. Although haze synthesis on a clear RS image is passable, our model trained with artificial hazy images suffers from unsatisfying performance on real haze. Alternatively, recent studies provide a new perspective that jointly employs ground-level and satellite images for various \textit{satellite-to-ground} cross-reference applications when either image source is insufficient \cite{shi2022beyond,lu2020geometry}.

\begin{figure*}[t]
	\centering
	\includegraphics[width=1\linewidth]{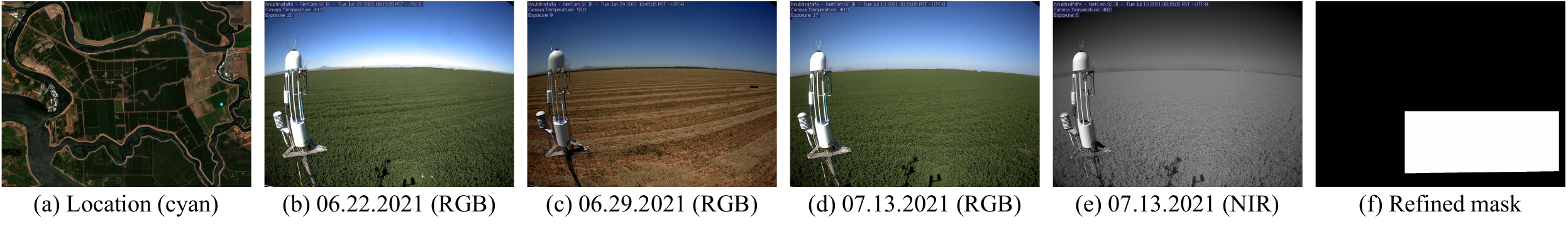}
	
	\vspace*{-1.4em}
	\caption{\small (a): Geographic location (cyan dot) of the PhenoCam observation site \texttt{bouldinalfalfa} \cite{phenocam2016} on an S2 image. (b)-(d): Example images of the ground-based imagery source, showing the visual appearance changes within only a one-month span. (e): Example near-infrared image on the same date of (d). (f): Refined mask based on the given one. \vspace*{-1.3em}}
	\label{fig:alfalfa}
\end{figure*}

Peak Signal-to-Noise Ratio (PSNR) and Structural Similarity Index (SSIM) are two commonly used metrics for
dehazing evaluation \cite{liu2021review,qin2020ffa,song2022vision},
however, they are hardly interpreted or deployed in practice as an indicator of the dehazing model upgrade. Our clients' feedback indicates that the increment in the value of PSNR or SSIM does not necessarily bring out quality improvement in human perception. Recent studies also confirm this misalignment in evaluating low-level CV tasks such as the super-resolution task \cite{reibman2006quality,metrics2021}. Therefore, designing a stereotype-breaking metric is necessary \cite{gu2016no,gu2013fisblim,gu2017model}.

To address the pain points in terms of the data and metrics above,
we propose a new evaluation metric for RS image dehazing assessment based on \emph{satellite-to-ground} multimedia image sources, called ``\underline{SA}tellite-to-\underline{G}round \underline{E}rror of \underline{NDVI}'' (\sage). Our key contributions are listed as follows.

\begin{itemize}
	\item To tackle the unavailability of a clear ground truth RS image, we leverage a sharp ground image dataset of vegetation phenology, which daily monitors multiple crop types worldwide using outdoor cameras. The error between the NDVI values of this clear ground image source and the dehazed RS image source contributes to the proposed \sage.
	\item Extensive experiments demonstrate that our objective \sage is capable of appropriately evaluating different dehazing models and more consistent with human perception than the conventional PSNR metric.
\end{itemize}

\section{Related Work}
\label{sec:related}
\textbf{Natural Image and Remote Sensing Image Dehazing.}
For natural images, traditional methods mainly utilized different priors for image restoration and improve the image contrast and saturation \cite{he2010single,zhu2015fast,galdran2018duality}. Recently deep learning-based methods aimed to generate a dehazed image from a hazy one \cite{cai2016dehazenet,qin2020ffa,song2022vision}. FFA-Net \cite{qin2020ffa} exploited channel and pixel attention modules to incorporate features on different channels to accomplish robust dehazing performance. DehazeFormer \cite{song2022vision} adopted reflection padding and modified normalization layers built upon Swin Transformer \cite{liu2021swin} to achieve state-of-the-art performance on various datasets. For RS scenarios, image dehazing is a rigid demand for various civilian purposes, like land planning and crop yield surveys \cite{liu2021review,kalra2022new,gu2019single}.

\textbf{Evaluation Metrics.}
Peak Signal-to-Noise Ratio (PSNR) and Structural Similarity Index (SSIM) are the two main objective metrics to evaluate the dehazing effects \cite{cai2016dehazenet,liu2021review,qin2020ffa,song2022vision}. The former gauges the pixel-wise error using Mean Squared Error (MSE), and the latter measures the structural similarity of pictures to estimate the quality of pictures after compression \cite{wang2004image}. However, they are not entirely consistent with human perception \cite{metrics2021} or even interpretable for clients who lack domain knowledge. In the context of our deployment scenario using RS imagery, our design of the proposed objective evaluation metric \sage benefits from the satellite-to-ground multimedia image reference sources and breaks the stereotype of the difficulty to measure low-level image tasks in RS images.

\section{Materials and Methods}
\label{sec:method}

\subsection{Data Acquisition and Pre-Processing}

\textbf{Satellite Image Data.} Instead of using Google Earth or other sources that are limited to academic use only, we utilize the Sentinel-2 (S2) satellite imagery with a 10m spatial resolution \cite{drusch2012sentinel}, which is \emph{commercial use free}, to construct our satellite image database. It is worth noting that our database pipeline prioritizes the clients' demands rather than this specific evaluation metric design. As required by our clients, the production of S2 images for a given time interval (e.g. every 8 days) must be guaranteed.
However, the original revisiting period of S2 satellites (around 4-5 days) is not fixed and an S2 image for each period is sometimes corrupted or defective, definitely conflicting with the clients' requirements. To this end, we exploit temporally adjacent images to composite an S2 image with a fixed time interval of every 8 days.
Fig.~\ref{fig:alfalfa}a provides an S2 image example from our satellite image database with the pinned geographic location of the observation site.

\textbf{Ground Image Data.} We leverage the PhenoCam database as our ground data source \cite{seyednasrollah2019tracking}. This database of monitoring vegetation phenology covers over 800 global observation locations using outdoor ground-based cameras. Due to the impact of diverse climates across worldwide locations on camera image availability, especially in rainy and snowy seasons, our focus is the cropland areas in the middle Central Valley of California, USA. The reasons are twofold: 1) this area enjoys a hot Mediterranean climate that rarely has snowy days; 2) the long-running camera deployment acquires abundant images. Specifically, we select an observation site that grows \emph{alfalfa} at Bouldin Island, CA \cite{phenocam2016}, where the image data is available from 2016 to the present. Since alfalfa is harvested several times a year, its vegetation appearance can change more frequently than other plants. This property is challenging but beneficial to our metric design and evaluation. Fig.~\ref{fig:alfalfa}b-\ref{fig:alfalfa}d exemplify the images captured at this site, showing the visual changes within only one month.

The deployed camera contains four RGBN channels: red, green, blue, and near-infrared (NIR) (shown in Fig.~\ref{fig:alfalfa}e). The frequency of image production of a camera is around every 30 minutes in the daytime every day. We refine the originally given binary mask to eliminate the region of the camera stand and the sky (shown in Fig.~\ref{fig:alfalfa}f). Thanks to our careful selection of the observation site and time span, the masked regions can be regarded as cloud-free and haze-free image sources, serving as a reliable reference for S2 image counterparts.

\setlength{\textfloatsep}{0pt}
\begin{algorithm}[ht]
    \caption{\sage}\label{alg:sage}
    \SetAlgoLined
    \KwIn{Satellite imgs: $\mathbf{I} = [\mathbf{I}_1, \mathbf{I}_2, \dots, \mathbf{I}_n]$; \newline Ground imgs: $\mathbf{G} = [\mathbf{G}_{1,1}, \mathbf{G}_{1,2}, \dots, \mathbf{G}_{l,t_l}]$; \newline NDVI threshold: $h$.}
    \KwFunc{\ding{202} $\psi$: Cloud detector; \ding{203} $\phi$: Dehazing; \newline \ding{204} $\lambda^{\mathbf{I}}$: NDVI for $\mathbf{I}$; \ding{205} $\Tilde{\lambda}^{\mathbf{G}}$: NDVI and post-proc for $\mathbf{G}$; \ding{206} $\mathtt{d}$: DTW.}
    \KwOut{$\mathrm{\overline{e}} \in \mathbb{R},~~\mathrm{\overline{e^{\phi}}} \in \mathbb{R}$.}
    \For{$i \leftarrow 1$ \KwTo $n$}{
    $\mathbf{M}^{\psi}_i \leftarrow \psi(\mathbf{I}_i)$\;
    $\mathbf{I}_i^{\psi} \leftarrow \mathbf{I}_i~\&~(\mathbf{1} - \mathbf{M}^{\psi}_i)$\;
    $\mathbf{I}_i^{\phi} \leftarrow \phi(\mathbf{I}_i^{\psi})$\;
    }
    $\mathbf{u} \leftarrow \lambda^{\mathbf{I}}(\mathbf{I}^{\psi});~~\mathbf{u}^{\phi} \leftarrow \lambda^{\mathbf{I}}(\mathbf{I}^{\phi});~~\mathbf{v} \leftarrow \Tilde{\lambda}^{\mathbf{G}}(\mathbf{G})$\;

    Normalize $\mathbf{u} \in \mathbb{R}^n$, $\mathbf{u}^{\phi} \in \mathbb{R}^n$, and $\mathbf{v} \in \mathbb{R}^m$\;

    $\mathbf{A} \leftarrow \mathtt{d}(\mathbf{u}, \mathbf{v}) \in \{0,1\}_{(n \times m)}$\;
    $\mathbf{A}^{\phi} \leftarrow \mathtt{d}(\mathbf{u}^{\phi}, \mathbf{v}) \in \{0,1\}_{(n \times m)}$\;

    Initialize $\mathrm{e}$, $\mathrm{e^{\phi}}$, and $k$ with $0$s\;

    \For{$i \leftarrow 1$ \KwTo $n$}{
    	
    \If{$|u_i - u_i^{\phi}| > h$}{    	
    $\mathbf{q} \leftarrow \{j|A_{ij} == 1; j \in \{1,\dots,m\}\}$\;
    $\mathbf{q}^{\phi} \leftarrow \{j|A^{\phi}_{ij} == 1; j \in \{1,\dots,m\}\}$\;
    $\mathrm{e} \leftarrow \mathrm{e} + \frac{1}{|\mathbf{q}|}\sum_{q \in \mathbf{q}} |u_i - v_q|$\;
    $\mathrm{e^{\phi}} \leftarrow \mathrm{e^{\phi}} + \frac{1}{|\mathbf{q}^{\phi}|}\sum_{q^{\phi} \in \mathbf{q}^{\phi}} |u_i^{\phi} - v_{q^{\phi}}|$\;
    $k \leftarrow k + 1$\;
    }
    }
    $\mathrm{\overline{e}} = \mathrm{e} / k;~~~~\mathrm{\overline{e^{\phi}}} = \mathrm{e^{\phi}} / k$\;
\end{algorithm}

\subsection{Design of \sage}

\textbf{Background of NDVI.}
The Normalized Difference Vegetation Index (NDVI) \cite{rouse1974monitoring} is an
technique to assess vegetation growth and condition by measuring the difference between NIR and red channels of an image, calculated as $\mathrm{NDVI} = (\mathrm{NIR} - \mathrm{Red}) / (\mathrm{NIR} + \mathrm{Red})$.
NDVI values range from -1 to +1. The higher values indicate more green and healthy vegetation and negative values often indicate water bodies. NDVI is widely used in environmental monitoring to assess vegetation cover, monitor changes in land use and land cover, and track the effects of climate change on ecosystems.

\textbf{Algorithm Details.}
Alg.~\ref{alg:sage} reveals our algorithm design of \sage.
Let $\mathbf{I} = [\mathbf{I}_1, \mathbf{I}_2, \dots, \mathbf{I}_n]$ denote the $n$-length sequence of 8-day composite S2 satellite images in one year, where $n$ is the number of valid satellite images. The ground image year-round sequence is $\mathbf{G} = [\mathbf{G}_{1,1}, \mathbf{G}_{1,2}, \dots, \mathbf{G}_{l,t_l}]$, where $l$ refers to the number of valid days and $t_l$ is the number of available images on the $l$-th day. For each $\mathbf{I}_i, i \in \{1, \dots, n\}$, a cloud detector $\psi(\cdot)$ is first used to void pixels occluded by cloud, and the remaining valid pixels are then dehazed by a dehazing model $\phi(\cdot)$.
The corresponding NDVI sequences ($\mathbf{u}$ and $\mathbf{u}^{\phi}$) of the cloud-cropped \textit{hazy} satellite images $\mathbf{I}^{\psi}$ and the \textit{dehazed} satellite images $\mathbf{I}^{\phi}$ are next calculated, respectively.

Note that the original NDVI values of ground images have a length of $l$, since they are calculated on a daily average basis. We then denoise them and conduct peak detection to remove minor fluctuation and maintain the key patterns, consequently achieving an $m$-length NDVI sequence $\mathbf{v}$ for ground images $\mathbf{G}$, where $m$ is the number of detected peaks and troughs.

Due to the inherent distance in the color space between the satellite imaging sensors and the ground camera sensors, we normalize the three NDVI sequences using a min-max scaling strategy.
Note that a \textit{hazy} NDVI value can be a salient outlier of a regular figure, hence applying the min-max scaling scheme on such time series data can cause serious offset errors. Those inappropriate offsets further damage the real relative distances between hazy $\mathbf{u}$ and dehazed $\mathbf{u}^{\phi}$. We normalize $\mathbf{u}$ using the parameters of $\mathbf{u}^{\phi}$ instead.

Next, we measure the similarity between the NDVI temporal sequences of satellite images and ground images. Owing to the different lengths ($n$~vs.~$m$) of NDVI values, the dynamic time warping (DTW) algorithm \cite{muller2007dynamic} is exploited to dynamically associate similar values of two time series with different temporal resolutions. We let $\mathbf{A}$ denote the binary adjacency matrix to present the DTW path. For the $i$-th of $n$ time stamps, we compare the difference between hazy $u_i$ and dehazed $u_i^{\phi}$. If the difference is greater than a given threshold, the dehazing effect is \emph{significant} at this time spot. We define $\mathrm{\overline{e}}$ and $\mathrm{\overline{e^{\phi}}}$ as the mean errors for hazy and dehazed NDVI values at all these significant time stamps, respectively.

\subsection{Use Description of \sage}
For the use of \sage, the hazy images of the PhenoCam observation site are provided as the gold standard reference dataset. To evaluate a dehazing model, we run the model on those hazy images and generate the dehazed ones. The dehazed images are used to calculate the values of $\mathrm{\overline{e^{\phi}}}$.

\section{Performance Evaluation}
\label{sec:result}

\begin{table}[t]
    \setlength{\tabcolsep}{4pt}
    \centering
    \caption{\small \sage results at the observation site \texttt{bouldinalfalfa} \cite{phenocam2016} from 2018 to 2022. \textit{Green}: difference between $\mathrm{\overline{e}}$ and $\mathrm{\overline{e^{\phi}}}$. \textit{Bold}: the better dehazing performance between FFA-Net and DehazeFormer. \vspace{-0.5em}}
    \vspace{-0.2em}
    \resizebox{\columnwidth}{!}{
        \begin{tabular}{cccc}
            \toprule
            \textit{Alfalfa} & $\mathrm{\overline{e}}$ ($\downarrow$) & $\mathrm{\overline{e^{\phi}}}$ ($\downarrow$) (FFANet\cite{qin2020ffa}) & $\mathrm{\overline{e^{\phi}}}$ ($\downarrow$) (DehazeFormer\cite{song2022vision}) \\
            \midrule
            2018             & 0.4457                                 & 0.2312\cgapg{-}{0.2145}                                                 & \textbf{0.1865}\cgaphl{-}{0.2592}                                                 \\
            2019             & 0.3259                                 & 0.3181\cgapg{-}{0.0077}                                                 & \textbf{0.3032}\cgaphl{-}{0.0227}                                                 \\
            2020             & 0.3190                                 & 0.2351\cgapg{-}{0.0839}                                                 & \textbf{0.2259}\cgaphl{-}{0.0931}                                                 \\
            2021             & 0.4385                                 & 0.2773\cgapg{-}{0.1612}                                                 & \textbf{0.2243}\cgaphl{-}{0.2142}                                                 \\
            2022             & 0.4370                                 & 0.2140\cgapg{-}{0.2230}                                                 & \textbf{0.1696}\cgaphl{-}{0.2674}                                                 \\
            \midrule
            Mean             & 0.3764                                 & 0.2719\cgapg{-}{0.1045}                                                 & \textbf{0.2219}\cgaphl{-}{0.1545}                                                 \\
            \bottomrule
        \end{tabular}
    }
    \vspace{0.7em}
    \label{tab:sage-alfalfa}
\end{table}

\subsection{Implementation Details}
For the satellite side, we collect 18,776 S2 image pairs of a hazy and a clean image at the proximate date. The Dark Channel Prior (DCP) \cite{he2010single} is used to determine whether one image is hazy or not. The default DCP threshold is set to 20. We choose the FFA-Net \cite{qin2020ffa} and the DehazeFormer-B \cite{song2022vision} since they demonstrated superior performance on various datasets in \cite{song2022vision} and they are deployed in our production environment. The images were cropped into patches with a size of 1024$\times$1024 and equipped with rotation and flip augmentations. We keep the default setting of both FFA-Net and Dehazeformer-B, respectively trained for 300 epochs. The threshold of NDVI difference $h$ is set to 0.1 by default.

For the ground side, we download the image data of this observation site spanning from 2018 to 2022. To avoid the camera being affected by the sunrise and sunset, we use the images from 11:00 to 13:00 when the sun is not in camera images, collecting approximately 4 images per day.

\subsection{Experiment Results}
\textbf{Evaluation of Dehazing Models Using \sage.}
Table~\ref{tab:sage-alfalfa} lists the evaluation results of dehazing models using \sage at the alfalfa observation site from 2018 to 2022. The significantly decreased error values $\mathrm{\overline{e^{\phi}}}$ compared with $\mathrm{\overline{e}}$ demonstrate the usability of our newly designed metric. Furthermore, the bold results indicate
the stable superiority of DehazeFormer-B over FFA-Net.
This observation is consistent with the results in \cite{song2022vision} that DehazeFormer-B overall outperformed FFA-Net on a remote sensing dataset and other datasets using the traditional PSNR and SSIM metrics.

\begin{table}[t]
    \setlength{\tabcolsep}{4pt}
    \caption{\small Averaged \sage on other observation sites \cite{phenocam2016}. \textit{Green}: difference between $\mathrm{\overline{e}}$ and $\mathrm{\overline{e^{\phi}}}$. \vspace{-0.7em}}
    \centering
    \resizebox{\columnwidth}{!}{
        \begin{tabular}{cccc}
            \toprule
            Mean          & $\mathrm{\overline{e}}$ ($\downarrow$) & $\mathrm{\overline{e^{\phi}}}$ ($\downarrow$) (FFANet\cite{qin2020ffa}) & $\mathrm{\overline{e^{\phi}}}$ ($\downarrow$) (DehazeFormer\cite{song2022vision}) \\
            \midrule
            \textit{Corn} & 0.3085                                 & 0.2374\cgapg{-}{0.0711}                                                 & \textbf{0.2285}\cgaphl{-}{0.0837}                                                 \\
            \textit{Rice} & 0.1602                                 & 0.1232\cgapg{-}{0.0370}                                                 & \textbf{0.1157}\cgaphl{-}{0.0445}                                                 \\
            \bottomrule
        \end{tabular}
    }
    \vspace{0.5em}
    \label{tab:sage-corn-rice}
\end{table}

\begin{figure}[t]
    \centering
    \includegraphics[width=1\linewidth]{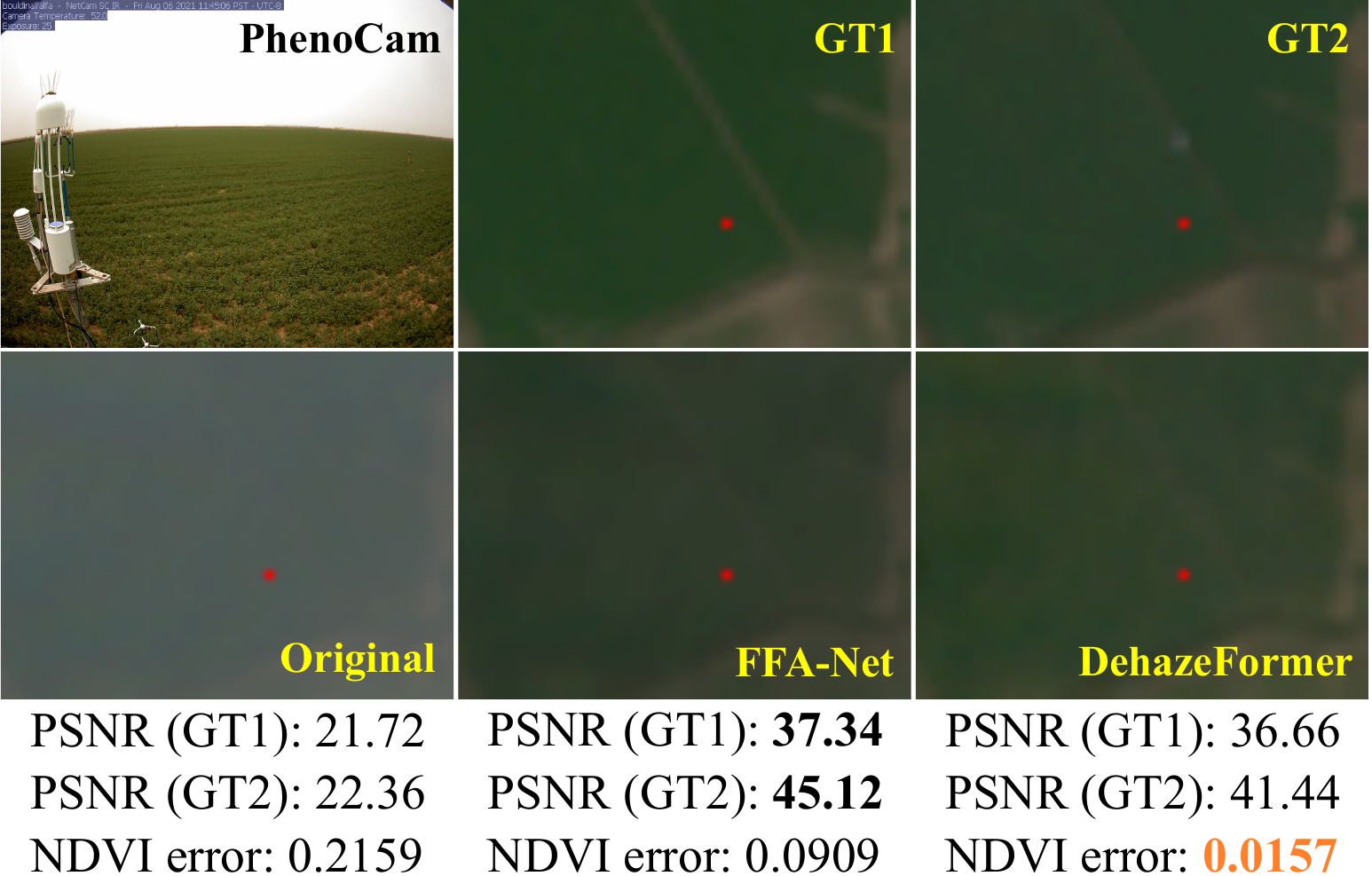}
    \vspace{-1.8em}
    \caption{\small Visualization of different PSNR and \sage results at the observation site \texttt{bouldinalfalfa} \cite{phenocam2016} (red dot). \textit{Top}: One clean ground image at a \emph{hazy} date and two clean satellite images at the \emph{adjacent} dates. \textit{Bottom-Left}: Original hazy image. \textit{Bottom-Center}: FFA-Net \cite{qin2020ffa}. \textit{Bottom-Right}: DehazeFormer \cite{song2022vision}. PSNR values and our new satellite-to-ground NDVI-based errors of three bottom images are provided. For better visualization, the brightness of all satellite images is increased with a fixed value. \vspace{0.5em}}
    \label{fig:results}
\end{figure}

\textbf{Comparison of \sage and PSNR.}
Although the general conclusions drawn from evaluating the dehazing models using either \sage or PSNR seem aligned as aforementioned, we next illustrate the shortcomings of PSNR. Fig.~\ref{fig:results} depicts the NDVI errors (the second addend on Line 15 and 16 in Alg.~\ref{alg:sage}) and the PSNR values calculated at the alfalfa observation site on a hazy date. Aside from the obviously poor PSNR value and large NDVI error of the original hazy image, we observe that even though PSNR values of FFA-Net are higher than those of DehazeFormer-B, the visualized results are not quite in accordance with the human perception. The image's color and texture reconstructed by DehazeFormer-B from the original hazy image are better. Instead, our objective NDVI error is able to match the subjective visual consistency better. We also conduct preliminary subjective evaluation from both our technical and client sides, and the results align with the \sage.

\textbf{\sage at Other PhenoCam Locations.}
We also test \sage at different PhenoCam locations, focusing on the crop types with few phenological periods than alfalfa, such as corn and rice \cite{phenocam2016}. The NDVI values at those locations do not have dramatic changes throughout the year. Table~\ref{tab:sage-corn-rice} shows similar observed relationships among $\mathrm{\overline{e}}$ and two $\mathrm{\overline{e^{\phi}}}$ values. However, due to the few phenological periods and available hazy dates, the differences in NDVI values are not noticeable as those at the alfalfa observation site.

\section{Conclusions}
\label{sec:concl}

In this paper, we leverage the satellite-to-ground philosophy to propose a new objective metric, \texttt{SAGE-NDVI}, for RS image dehazing evaluation. A public phenology observation resource containing ground images is exploited to calculate the vegetation index error between RS and ground images at a hazy date.
Our metric's capability of appropriately evaluating various dehazing models and conforming to human visual perception is demonstrated by extensive experiments.

\bibliographystyle{IEEEtran}
\bibliography{MAIN-CR}

\end{document}